\documentclass{article}

    \PassOptionsToPackage{numbers, compress}{natbib}

    \usepackage[preprint]{neurips_2024}

\usepackage[utf8]{inputenc} 
\usepackage[T1]{fontenc}    
\usepackage{hyperref}       
\usepackage{url}            
\usepackage{booktabs}       
\usepackage{amsfonts}       
\usepackage{nicefrac}       
\usepackage{microtype}      
\usepackage{xcolor}         
\usepackage{amsmath}
\usepackage{graphicx}
\usepackage{bm}

\title{Eliciting the Priors of Large Language Models \\ using Iterated In-Context Learning}

\author{%
  Jian-Qiao Zhu\\
  Department of Computer Science\\
  Princeton University\\
  \texttt{jz5204@princeton.edu} \\
  \And
  Thomas L. Griffiths\\
  Department of Psychology and Computer Science\\
  Princeton University\\
  \texttt{tomg@princeton.edu} \\
}

\begin{document}

\maketitle

\begin{abstract}
  As Large Language Models (LLMs) are increasingly deployed in real-world settings, understanding the knowledge they implicitly use when making decisions is critical. One way to capture this knowledge is in the form of Bayesian prior distributions. We develop a prompt-based workflow for eliciting prior distributions from LLMs. Our approach is based on iterated learning, a Markov chain Monte Carlo method in which successive inferences are chained in a way that supports sampling from the prior distribution.
  We validated our method in settings where iterated learning has previously been used to estimate the priors of  human participants -- causal learning, proportion estimation, and predicting everyday quantities. We found that priors elicited from GPT-4 qualitatively align with human priors in these settings. We then used the same method to elicit priors from GPT-4 for a variety of speculative events, such as the timing of the development of superhuman AI. 
\end{abstract}

\section{Introduction}
As Large Language Models (LLMs) become increasingly integrated into diverse real-world applications, there is a pressing need to understand their decision-making  processes \cite{bengio2023managing}. This is particularly crucial in scenarios where LLMs are granted agency to act and make decisions independently \cite{yao2022react, wu2023autogen}. Such agentic applications can significantly affect outcomes in sectors such as healthcare \cite[e.g.,][]{nori2023capabilities}, finance \cite[e.g.,][]{wu2023bloomberggpt}, and legal services \cite[e.g.,][]{katz2024gpt}, where the implications of errors or biased decisions can be profound. An essential component in understanding the decision-making processes of LLM is identifying the background knowledge they implicitly possess. For instance, consider a scenario where LLMs are asked to estimate a person’s lifespan (i.e., hypotheses about the person's lifespan, $h$) based on a description of their current status (i.e., data about the person, $d$). Does the estimate produced by the LLM depend exclusively on the information provided in the description, or is it also shaped by background knowledge concerning the typical lifespan of individuals?

\begin{figure}[t!]
    \centering
    \includegraphics[width=\textwidth]{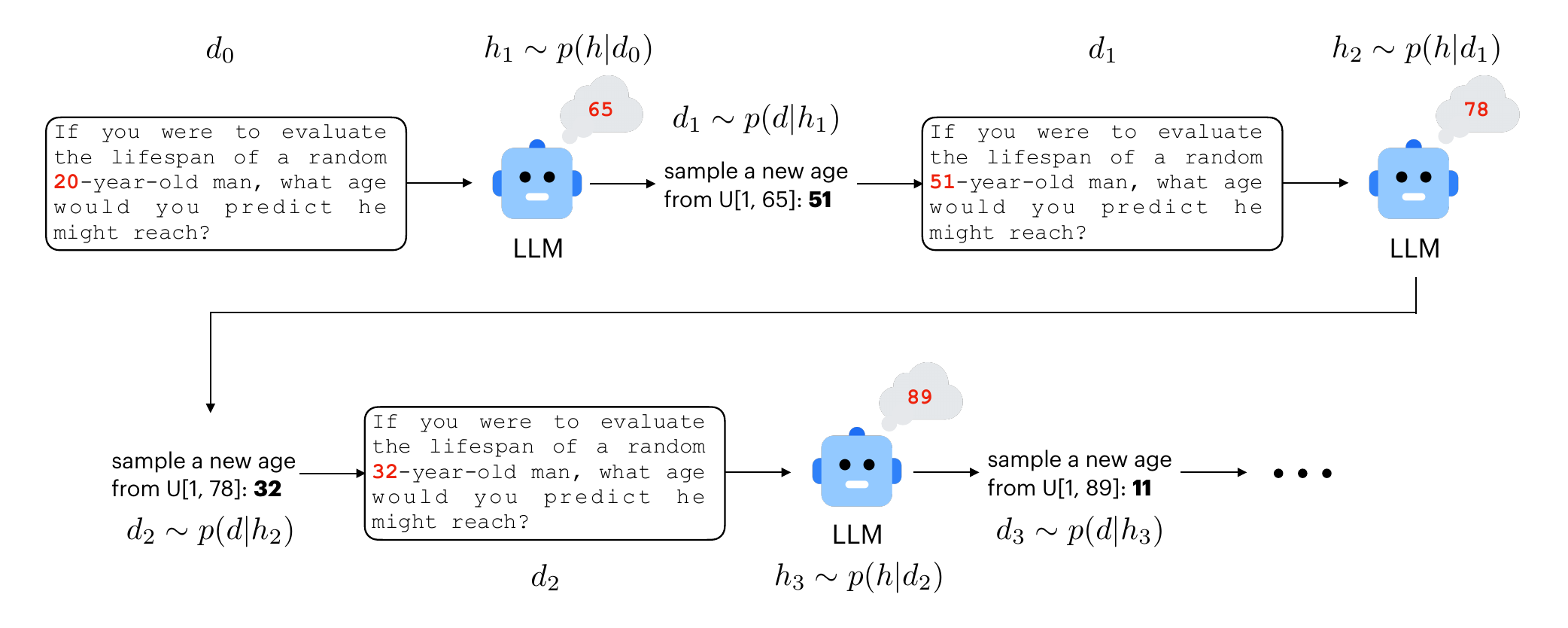}
    \caption{Illustration of an iterated in-context learning procedure to elicit the implicit prior of an LLM regarding male lifespan. At each iteration, the LLM is given the current age of a random man and is prompted to predict the individual’s remaining lifespan. This predicted lifespan is then used to generate a new current age for the next iteration. The new age is a random sample from the uniform distribution between 1 and the predicted lifespan. This implements a Markov chain Monte Carlo algorithm for sampling from the prior $p(h)$.}
    \label{fig:method_illustration}
\end{figure}

To explore this question we adopt a Bayesian perspective, formalizing this background knowledge as a prior distribution over the hypothesis space (i.e., $p(h)$) \cite{griffiths2010probabilistic, tenenbaum2001generalization}. This approach enables us to assess, in probabilistic terms, how such prior knowledge affects the judgments and decisions made by LLMs, thereby enhancing our understanding of their underlying decision-making mechanisms. To elicit the priors of LLMs, we draw inspiration from cognitive psychology and develop an iterated learning procedure, a Markov chain Monte Carlo (MCMC) method. As illustrated in Figure \ref{fig:method_illustration} for the lifespan example, this method involves using successive inferences from LLMs in a sequential manner that supports direct sampling from the prior distribution, mirroring techniques used in psychological studies to elicit human priors \cite[e.g.,][]{griffiths2006revealing, reali2009evolution, kalish2007iterated}.

Given the established efficacy of iterated learning in eliciting human priors in cognitive psychology, we explored its applicability to LLMs. We conducted experiments using tasks from three distinct domains -- estimations of causal strengths, proportion, and everyday quantities -- where human priors are well-documented \cite{reali2009evolution, yeung2015identifying, lewandowsky2009wisdom}. These experiments successfully elicited priors from GPT-4. The priors recovered from GPT-4 not only align closely with human priors but can surpass the performance of generic priors, such as a uniform prior, in explaining decisions made by GPT-4 in these settings.

Encouraged by the empirical evidence demonstrating shared priors between GPT-4 and humans across a broad range of tasks, we further investigated the potential of iterated learning to uncover priors in settings where priors are challenging to estimate directly from LLMs using standard prompting techniques. To illustrate, we applied iterated learning to elicit GPT-4’s priors for three speculative events: the advent of superhuman AI, the achievement of zero carbon emissions, and the establishment of a Mars colony. The distributions recovered from GPT-4 suggest the model has plausible priors for these speculative events.

\section{Background}

\textbf{Iterated learning in cognitive science.}
Iterated learning was first introduced as a model for language evolution \cite{kirby2001spontaneous}. Language evolution can be conceptualized as the process through which languages are transmitted across successive generations of learners. In this model, an initial learner observes a corpus of linguistic data (for example, a collection of utterances), formulates a hypothesis about the underlying language that produced these utterances, and subsequently produces a new set of utterances. These are then used as data for the next learner in the sequence. Research has demonstrated that generational pressure on language transmission fosters the emergence of compositionality, realistic patterns of language dynamics, and several other observed properties of natural languages \cite{kirby2001spontaneous, christiansen2003language}.

Motivated by these results, an analysis of iterated learning with Bayesian learners showed that such a process will converge towards the prior distribution assumed by the learners \cite{griffiths2007language}.
The analysis assumes that all learners share the same prior distribution over hypotheses $p(h)$ and likelihood function $p(d|h)$, which indicates how probable it would be to see data $d$ if $h$ were true. Each learner sees data $d$ generated by the previous learner, samples a hypothesis $h$ from $p(h|d) \propto p(d|h)p(h)$, and then generates data for the next learner from $p(d|h)$. This process implements a Gibbs sampler for the joint distribution $p(d,h) = p(d|h)p(h)$, a form of Markov chain Monte Carlo. The stationary distribution on hypotheses is thus the prior $p(h)$, and samples from the prior can be drawn by running the iterated learning process long enough to converge to this distribution.

These theoretical results suggest that implementing iterated learning in the laboratory may be an effective way to identify the prior distributions of human learners \cite{kalish2007iterated}. Based on the connection to Gibbs sampling, doing so involves implementing a form of Markov chain Monte Carlo with people \cite{sanborn2007markov}. Researchers have indeed successfully used iterated learning to elicit human priors for various kinds of cognitive phenomena, such as concepts \cite{griffiths2008using}, categories \cite{canini2014revealing}, causal relationships \cite{yeung2015identifying}, proportions \cite{reali2009evolution}, and everyday quantities \cite{lewandowsky2009wisdom}.

\textbf{Iterated learning in deep learning.}
The \textit{learning bottleneck} in iterated learning limits the amount of information that can be transmitted and acquired by successive generations of learners, creating a selection pressure for communicative efficacy that promotes compositionality \cite{kirby2001spontaneous}. These advantages have led deep learning researchers to use iterated learning to encourage the development of structure and compositionality in neural networks \cite{vani2021iterated, rajeswar2022multi, ren2024improving}. However, neural networks are still far from being idealized Bayesian learners, as assumed in the limit of iterated learning presented above. To bridge the gap between neural networks and Bayesian learners, seeded iterated learning methods have been proposed \cite{lu2020countering, lu2020supervised}. Instead of using a randomly initialized or pretrained neural network, seeded iterated learning methods use duplicated student and teacher networks at each iteration. Experimental results demonstrate that seeded iterated learning is effective in preserving the initial language structure acquired through pretraining while adapting the model to domain-specific tasks \cite{lu2020countering}. However, neglecting the gap between neural networks and Bayesian learners risks turning the iterated learning process into a degenerative one, where models gradually forget the true underlying data distribution -- a phenomenon known as \textit{model collapse} \cite{shumailov2023curse}. This issue has been frequently observed in machine learning, where training models on data produced by other models leads to forgetting improbable events, causing irreversible defects in the resultant models \cite{aljundi2019task, shumailov2023curse}.

\section{Eliciting Priors from GPT-4 using Iterated In-Context Learning}

By contrast to previous work on iterated learning in machine learning, we use iterated learning applied to {\em in-context} learning rather than {\em in-weight} learning. That is, while previous work has focused on iterating the process of training a neural network's weights, we focus on generating predictions from a neural network that has already been pre-trained and has fixed weights. In this setting, we are relying on the network's ability to generate responses to prompts using that fixed set of weights. In doing so, we can capture the implicit knowledge encoded in those weights in the resulting prior distribution. This implementation of iterated learning also avoids the model collapse problem because there is no explicit training. 

This approach assumes that it is reasonable to interpret in-context learning as a form of Bayesian inference. Fortunately, a number of recent papers have provided support for this idea \cite{xie2021explanation, zhang2023deep}. We thus hypothesize that the theoretical results for iterated learning with Bayesian agents are applicable to LLMs. Specifically, in-context iterated learning with LLMs should converge to responses that support sampling from the prior distribution.

To test the hypothesis that iterated in-context learning should reveal the prior distributions of LLMs, we incorporated GPT-4 \cite{achiam2023gpt} into a prompt-based iterated learning procedure. At each iteration $t$, GPT-4 undertakes a prediction task using the data $d_{t-1}$. The model's prediction is recorded as $h_t$. Subsequently, we employ generic likelihood functions that are a reasonable match for the sampling process producing the described data to randomly generate the data for the next iteration, $d_t \sim p(d|h_t)$. For instance, we applied the method depicted in Figure \ref{fig:method_illustration} to investigate GPT-4's prior beliefs about men's lifespans. In this procedure, the LLM is prompted to estimate the lifespan of a random man, given information about his current age. The age of the man encountered in the next prompt is then uniformly sampled from a range extending from 1 to the lifespan predicted in the previous iteration, matching the probability of randomly encountering the man at this point in his life. By iteratively applying this procedure, we expect the final prediction made by GPT-4 will converge on a stationary distribution that reflects the model's prior beliefs about human life expectancy. 
 
In the experiments presented in the remainder of the paper, 100 iterated learning chains were implemented with random seeds. We conducted 12 iterations for each chain. The temperature of GPT-4 was fixed at 1, consistent with the idea of sampling from the posterior.

\section{Iterated Learning in Settings with Known Human Priors}

To determine whether GPT-4’s implicit priors resemble human priors, we first elicited GPT-4’s implicit priors using a set of iterated learning tasks that have previously been used to infer human priors (see Table \ref{tab:IL_with_known_human_priors}).

\begin{table}[h!]
    \centering
    \caption{Overview of human priors elicited using the iterated learning method. }
    \begin{tabular}{llll}\toprule
        Chain & Seeds & Likelihood functions & Trials  \\ \hline
        Generative causal strengths & $w_0=\{0.3, 0.7\}, w_1=\{0.3, 0.7\}$ & noisy-OR & 7 \\
        Preventive causal strengths & $w_0=\{0.3, 0.7\}, w_1=\{0.3, 0.7\}$ & noisy-AND-NOT & 8 \\
        Coin flips & $p(\text{Head})=\{0.3, 0.5, 0.7\}$ & $\text{Bin}(10, h_{t-1})$ & 1 \\
        Lifespan (male) & $t_\text{max}=150$ years old & $U[1, h_{t-1}]$ & 2\\
        Movie grosses & $x_\text{max}=3000$ million dollars & $U[0,h_{t-1}]$ & 11\\
        Length of poems & $x_\text{max}=200$ lines & $U[1,h_{t-1}]$  & 10 \\
        Reign of Pharaohs & $t_\text{max}=100$ years & $U[0, h_{t-1}]$ & 8 \\
        Movie run times & $t_\text{max}=800$ minutes & $U[0, h_{t-1}]$ & 5 \\
        Cake baking times & $t_\text{max}=120$ minutes & $U[0, h_{t-1}]$ & 3\\
         \bottomrule
    \end{tabular}\\
    \textit{Note.} Trials column indicates the estimated number of trials to convergence. Seeds determine the generation of initial data ($d_0$).
    \label{tab:IL_with_known_human_priors}
\end{table}

\subsection{Causal strength}

\begin{figure}[h!]
    \centering
    \includegraphics[width=\textwidth]{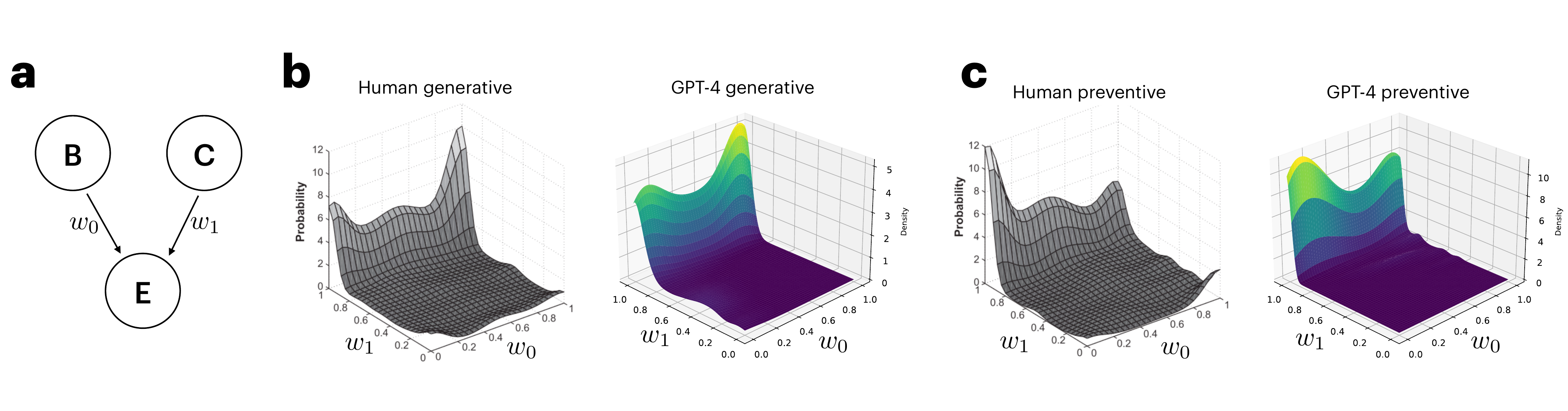}
    \caption{\textbf{Priors on causal strengths.}
    \textbf{(a)} The causal graphical model.
    \textbf{(b)} Smoothed empirical estimates of human (left) and GPT-4 (right) priors on causal strength produced by iterated learning for generative cases. 
    \textbf{(c)} Smoothed empirical estimates of human (left) and GPT-4 (right) priors on causal strength produced by iterated learning for preventive cases.
    Human data in panel (b) and (c) were adapted from \cite{yeung2015identifying}.}
    \label{fig:causal_prior}
\end{figure}

To ensure accuracy and clarity in the use of LLMs for causal inference, it is crucial to understand the implicit priors about causal relationships embedded within these models. We examined an elemental problem of causal induction \cite{griffiths2005structure} involving two potential causes and one effect (see Figure \ref{fig:causal_prior}a). In this model, the causal system is represented by three variables: the background cause (B), the candidate cause (C), and the effect (E). Both B and C can independently cause E, and this relationship is depicted by edges directed from both B and C to E. The causal strengths of B and C are represented by $w_0$ and $w_1$, respectively.

We further assume that B is always present and generative, meaning it consistently increases the probability of E. However, C can be either generative or preventive. In the generative scenario, either B or C can cause E; in the preventive scenario, only B can cause E, while C may inhibit E. Additionally, E cannot occur unless it is caused by either B or C. Depending on the functional form of the causal relationships, the probability of observing an effect given two causes is expressed differently: a noisy-OR likelihood function is used for generative causes and a noisy-AND-NOT likelihood function for preventive causes \cite{cheng1997covariation, griffiths2005structure, pearl2009causal}. The noisy-OR function gives the probability of observing E as:
\begin{align}
    p(e^+ | C^+) & = 1-(1-w_0)(1-w_1), \text{~~if C is present} \\
    p(e^+ | C^-) & = 1-(1-w_0), \text{~~if C is absent}
\end{align}
Noisy-AND-NOT gives the probability of observing E as:
\begin{align}
    p(e^+ | C^+) & = w_0(1-w_1), \text{~~if C is present} \\
    p(e^+ | C^-) & = w_0, \text{~~if C is absent}
\end{align}

Here, we are particularly interested in the prior distribution on causal strengths implicitly used by LLMs: $p(w_0, w_1)$. One potential prior is the uniform prior, arguably the simplest non-informative prior, which assigns equal probability to all possible values of $w_0$ and $w_1$ \cite{jaynes2003probability, griffiths2005structure}. Another prior that LLMs might employ is the \textit{sparse and strong} prior, which is motivated by simplicity principles suggesting that people favor necessary and sufficient causes without complex interactions \cite{lu2008bayesian}. The sparse and strong prior is defined as follows:
\begin{align}
    p(w_0, w_1) & \propto e^{-\alpha(w_0+1-w_1)}+e^{-\alpha(1-w_0+w_1)} , \text{~~for the generative case} \\
    p(w_0, w_1) & \propto e^{-\alpha(1-w_0+1-w_1)}+e^{-\alpha(1-w_0+w_1)} , \text{~~for the preventive case}
\end{align}
where $\alpha$ is a free parameter representing the strength of belief in the sparsity and strength of causes. When $\alpha=0$, the sparse and strong priors become identical to a uniform prior. Based on previous parameter estimation from human data, we also fixed  $\alpha=5$ \cite{lu2008bayesian}. Finally, LLMs could also employ an empirical prior that delineates a specific relationship between  $w_0$  and  $w_1$  but does not possess a precise mathematical description.

To elicit the empirical prior on causal strengths from LLMs, we implemented an iterated learning procedure with GPT-4 based on an experiment conducted with human participants \cite{yeung2015identifying}. The prompts used a cover story involving the influence of various proteins on gene expression (see Appendix \ref{ap:prompts} for details). The iterated learning chain was initiated with four possible pairs of $(w_0, w_1)$: $(0.3, 0.3)$, $(0.3, 0.7)$, $(0.7, 0.3)$, $(0.7, 0.7)$. The number of DNA fragments exposed and not exposed to the protein was fixed at 16 each (i.e., $N(C^+)=N(C^-)=16$). At each iteration $t$, we elicited GPT-4’s estimates of the causal strengths: $p(w_0, w_1 | d_{t-1})$. The data presented at iteration $t$ was a random sample drawn from the likelihood function based on GPT-4’s estimates from the previous iteration. 

Each chain consisted of 12 iterations and was randomly initialized 25 times for each of the 4 seeds, resulting in a total of 100 chains. Using a Mann-Whitney U test with a significance level of $p<.05$, we found that the chains converged to a stationary distribution by iterations 7 and 8 for generative and preventive causal strengths, respectively. The empirical distributions of $w_0, w_1$ at iteration 12, smoothed with a Gaussian kernel, were then considered the empirical prior of causal strengths (see Figure \ref{fig:causal_prior}b and Figure \ref{fig:causal_prior}c). The empirical priors derived from GPT-4 closely resemble those observed in human experiments \cite{yeung2015identifying}. Since human data are only available in the form of visualizations provided in the paper, we rely on visual comparisons to analyze the priors of both humans and GPT-4.

To further investigate which prior better captures GPT-4’s decisions about causal relationships, we elicited an additional set of causal judgments from GPT-4. We used the same cover story as in iterated learning but varied the number of DNA fragments exposed and not exposed to the protein, which was previously fixed at 16. Now $N(C^+)$ and $N(C^-)$ could take values from 8, 16, and 32, leading to a total of $3\times 3=9$ possible combinations of sample sizes. When the sample size is 8, $N(e^+)$ takes all possible integer values from 0 to 9; when the sample size is 16, $N(e^+)$ takes integer values in increments of 2; and when the sample size is 32, $N(e^+)$ takes integer values in increments of 4. This results in each sample size contributing 9 data points to the causal judgments.

To explain GPT-4’s causal judgments, we developed three Bayesian models based on those used to model human causal judgments \cite{griffiths2005structure,lu2008bayesian,yeung2015identifying}. Each model assumed a different prior. The posterior distribution was obtained by multiplying the prior of causal strength with the appropriate likelihood for the causal direction (i.e., generative or preventive):
\begin{align}
    p(w_0, w_1 |d) \propto p(w_0, w_1)p(d|w_0, w_1)
\end{align}
For all three Bayesian models, numerical methods were employed, discretizing the space of $w_0,w_1$ into a grid of $101\times101$ points. The mean of the posterior distribution was taken as the Bayesian model's prediction. We then compared the posterior means to the causal judgments produced by GPT-4. The results, summarized in Table \ref{tab:compare_causal_priors}, indicate that the empirical prior outperformed the uniform prior and the sparse and strong prior in all except the preventive case when measured by RMSD (see Appendix \ref{ap:bayes_model_predict_gpt_responses} for detail). These results suggest that we have successfully recovered the implicit prior of causal strengths using iterative learning with GPT-4.

\begin{table}[h!]
    \centering
    \caption{Comparison of Bayesian models of causal induction with various priors and GPT-4’s causal judgments using Pearson's $r$ and root-mean-squared deviation (RMSD).}
    \begin{tabular}{ccccc}
    \toprule
       Causal direction & Metric & Uniform prior & Sparse and strong prior & Empirical prior \\ \hline
       Generative & Pearson's $r$ ($\uparrow$) &  $0.85$ & $0.79$ & $\bm{0.86}$ \\
       & RMSD ($\downarrow$) &  $0.21$ & $0.25$ & $\bm{0.19}$ \\
       Preventive & Pearson's $r$ ($\uparrow$) & $0.72$ & $0.68$ & $\bm{0.79}$\\
       & RMSD ($\downarrow$) &  $\bm{0.26}$ & $0.29$ & $0.27$ \\
       \bottomrule
    \end{tabular}
    \label{tab:compare_causal_priors}
\end{table}

\begin{figure}[h!]
    \centering
    \includegraphics[width=\textwidth]{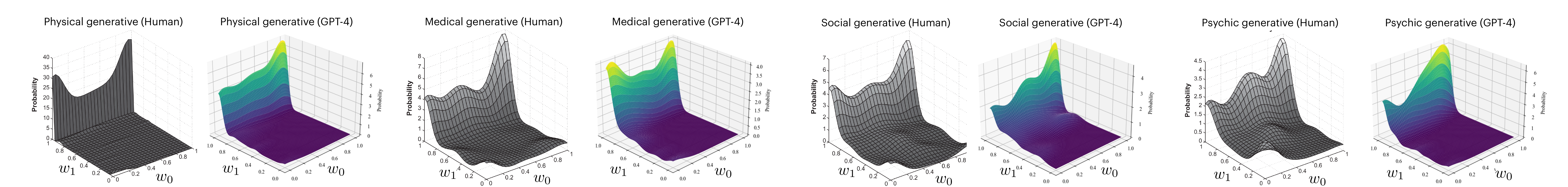}
    \caption{Comparison of causal generative priors using alternative cover stories between humans and GPT-4. Human data adapted from \cite{yeung2015identifying}. Detailed prompts are provided in Appendix \ref{ap:prompts}.}
    \label{fig:other_causal_priors}
\end{figure}

Since our primary interest is in comparing priors for causal directions (generative vs. preventive), we focused on the gene/protein cover story, as it is the only one that includes human priors for both generative and preventive causal directions. However, there are four other cover stories with human priors on the generative case \cite{yeung2015identifying}.We conducted four additional iterated in-context learning experiments with GPT-4, using alternative cover stories along with the same seeds and likelihood functions. As shown in Figure \ref{fig:other_causal_priors}, the implicit priors recovered from GPT-4 also align with the human priors.

\subsection{Proportion estimation}

Another setting with known human priors from iterated learning is proportion estimation \cite{reali2009evolution, zhu2020bayesian}. In these studies, human participants were asked at each iteration to judge the frequency of a binary event, such as a coin flip or a choice between two words \cite{reali2009evolution}. We implemented an iterated learning chain with GPT-4 to replicate this process using the cover story of coin flips. At each iteration, GPT-4 received the outcomes of 10 random coin flips, generated based on the previous iteration’s $p(\text{Head})$:
\begin{align}
    N(\text{Head}) \sim \text{Bin}(10, p(\text{Head}))
\end{align}
Then, GPT-4 was asked to report a new $p(\text{Head})$ by imagining throwing the same coin another 100 times. This reported $p(\text{Head})$  was then used to generate coin flip data for the next iteration (see Appendix \ref{ap:prompts} for detailed prompts).

\begin{figure}[h!]
    \centering
    \includegraphics[width=0.95\textwidth]{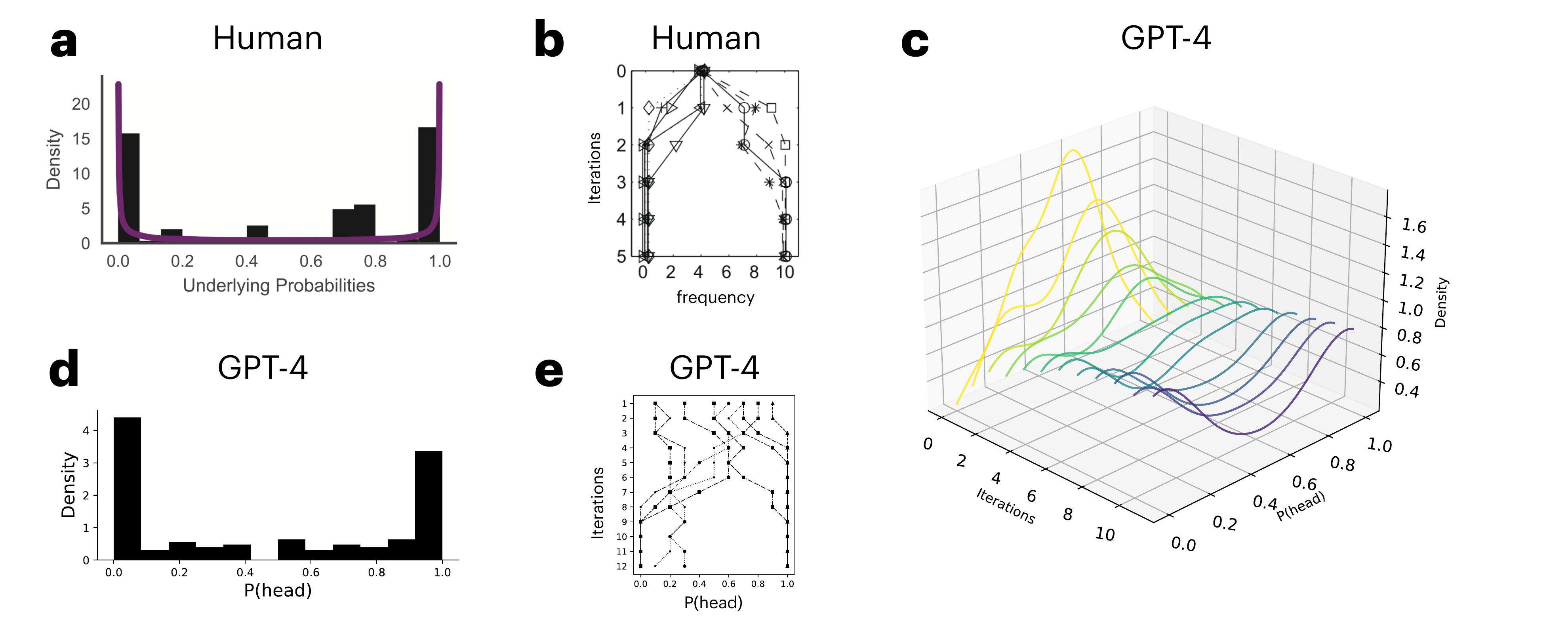}
    \caption{\textbf{Priors on proportion estimation.}
    \textbf{(a)} The empirical distribution of probability-describing phrases from the British National Corpus. Figure adapted from \cite{zhu2020bayesian}.
    \textbf{(b)} Example iterated learning chains for human participants estimating the proportion of binary events. Figure adapted from \cite{reali2009evolution}. 
    \textbf{(c)} The evolution of GPT-4’s estimation of binary events using iterated learning.
    \textbf{(d)} The histogram of GPT-4’s proportion estimation in the final (12th) iteration.
    \textbf{(e)} Example iterated learning chains for GPT-4 estimating the proportion of binary events, for comparison with human data.}
    \label{fig:frequency_priors}
\end{figure}

The evolution of the distribution of $p(\text{Head})$ is shown in Figure \ref{fig:frequency_priors}c. The distribution gradually shifted towards a U-shaped distribution, with most of its mass close to 0 or 1 (see Figure \ref{fig:frequency_priors}d for the final iteration's histogram and Figure \ref{fig:frequency_priors}e for a few example chains). 
The recovered prior of GPT-4 matches human priors (see Figure \ref{fig:frequency_priors}a). Moreover, the chain evolution from GPT-4 aligns with patterns observed in human data for iterated learning of the proportions of two words used as labels for an object (see Figure \ref{fig:frequency_priors}b).
Due to the absence of human data beyond the visualizations presented in the papers, we depend on visual comparisons to examine the priors of humans and GPT-4. Given that in these settings the prior has only a weak effect on proportion estimation (which motivated the use of iterated learning to study human priors in this context), we did not further test different priors using Bayesian models.

\subsection{Everyday quantities.}

A third class of tasks with known human priors elicited by iterated learning methods concerns everyday quantities \cite{lewandowsky2009wisdom}. These tasks can be broadly summarized as future-prediction tasks, where participants repeatedly provided predictions for a quantity $t_\text{future}$ in response to a given value of $t_\text{present}$. In our example of estimating a man's lifespan, $t_\text{future}$ would be the total lifespan and $t_\text{present}$ the age at which the man was encountered. Typically, the probe value of $t_\text{present}$ is randomly sampled from an interval ranging between 0 and the previous $t_\text{future}$:    $t_\text{present} \sim U[0, t_\text{future})$.

\begin{figure}[h!]
    \centering
    \includegraphics[width=\textwidth]{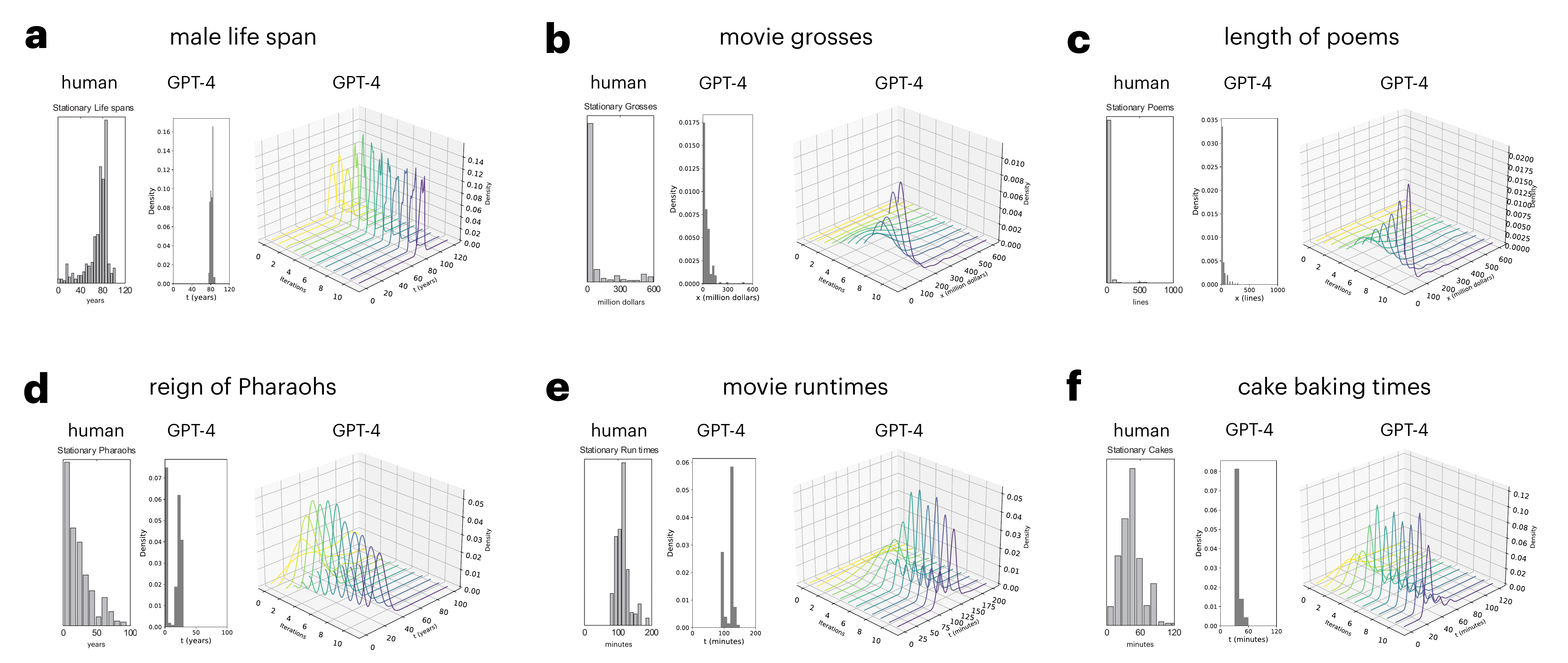}
    \caption{\textbf{Priors on everyday quantities.} Each panel displays the elicited prior using iterated learning on human participants (left), the histogram of GPT-4’s final iteration of predictions (middle), and the evolution of GPT-4’s predictions across iterated learning iterations (right). Human data adapted from \cite{lewandowsky2009wisdom}.}
    \label{fig:priors_on_everyday_quantities}
\end{figure}

We implemented all six everyday quantities tested in \cite{lewandowsky2009wisdom} ranging from male lifespan, movie grosses, length of poems, reign of Pharaohs, movie runtimes, and cake baking times (see Figure \ref{fig:priors_on_everyday_quantities}a-f). Because the likelihood function is a uniform distribution, meaning that the posterior distribution depends solely on the prior, we did not compare Bayesian models with different priors. Instead, we focused on directly comparing the recovered priors from GPT-4 to those from human participants. As shown in Figures \ref{fig:priors_on_everyday_quantities}(a-f), the modes of the priors from humans and GPT-4 were matched. Because human data are limited to the figures included in the paper, we use visual comparisons to evaluate the priors of humans and GPT-4. However, the overall distribution differed sometimes, especially for the Pharaohs. This is actually a case where people's beliefs are systematically incorrect -- by applying modern expectations about human lifespans to Ancient Egypt, people overestimate the length of the reigns of Pharaohs \cite{griffiths2006optimal}. GPT-4 produces more appropriate predictions in this setting.


\section{Iterated Learning as a Method for Estimating a Wider Range of Priors}

Applying iterated learning to estimate the priors of GPT-4 on causal strengths, proportion estimation, and everyday quantities reveals qualitative similarities with human priors. This suggests that LLMs have successfully learned human-like priors when producing judgments and predictions. Motivated by these findings, we aimed to test some speculative events that (i) have no known human priors, (ii) are difficult to quantify directly through prompts to GPT-4, and (iii) lack explicit consensus among humans. Iterated learning might serve as a unique way to address these three challenges because the priors recovered from LLMs using iterated learning are likely to resemble the implicit priors that people assume but have not yet explicitly manifested.

In principle, iterated learning is broadly applicable to a wide range of speculative events. However, LLMs typically avoid speculating on future events involving sensitive or potentially harmful topics (see Appendix \ref{ap:censored_speculative_events} for examples). These topics include political outcomes (e.g., predicting the winner of the U.S. presidential election in 2024), market forecasts (e.g., forecasting the price of Bitcoin in December of this year), personal futures (e.g., determining the likelihood of obtaining a recently interviewed job), legal outcomes (e.g., the outcome of ongoing investigations into public figures like Donald Trump), technological breakthroughs (e.g., the discovery of a cure for cancer next year), disasters (e.g., predicting the timing of the next earthquake in California), and specific dates for future events (e.g., when self-driving cars will become the primary mode of transportation worldwide). Generally, LLMs are restricted from making definitive predictions on sensitive and impactful issues related to speculative future events.

\begin{table}[h!]
    \centering
    \caption{Overview of GPT-4's priors on speculative events elicited using the iterated learning method.}
    \begin{tabular}{llll}\toprule
        Chain & Seeds & Likelihood functions & Trials  \\ \hline
        Superhuman AI & $t_\text{max}=2200$ year & $U[2024, h_{t-1}]$ & 5\\
        Zero carbon emission & $t_\text{max}=2200$ year & $U[2024, h_{t-1}]$ & 5 \\   
        Mars colony & $t_\text{max}=2200$ year & $U[2024, h_{t-1}]$ & 9 \\
         \bottomrule
    \end{tabular}\\
    \textit{Note.} Trials column indicates the estimated number of trials to convergence. Seeds determine the generation of initial data ($d_0$).
    \label{tab:IL_with_unknown_human_priors}
\end{table}

To illustrate the utility of eliciting priors from LLMs using iterated in-context learning rather than direct prompting, we focus on three technology and climate-related events: (i) the timing of the development of superhuman AI, (ii) the timing of achieving zero carbon emissions, and (iii) the timing of establishing a Mars colony. These events are particularly well-suited to our existing framework because they involve a clear two-stage completion process, similar to the future-prediction tasks illustrated in Figure \ref{fig:priors_on_everyday_quantities}. For example, superhuman AI can only be achieved if human-level AI has already been realized. Similarly, zero carbon emissions are possible only if the majority of energy use is renewable, and establishing a Mars colony is typically contingent upon the prior establishment of a Moon colony.

\begin{figure}[t!]
    \centering
    \includegraphics[width=0.9\textwidth]{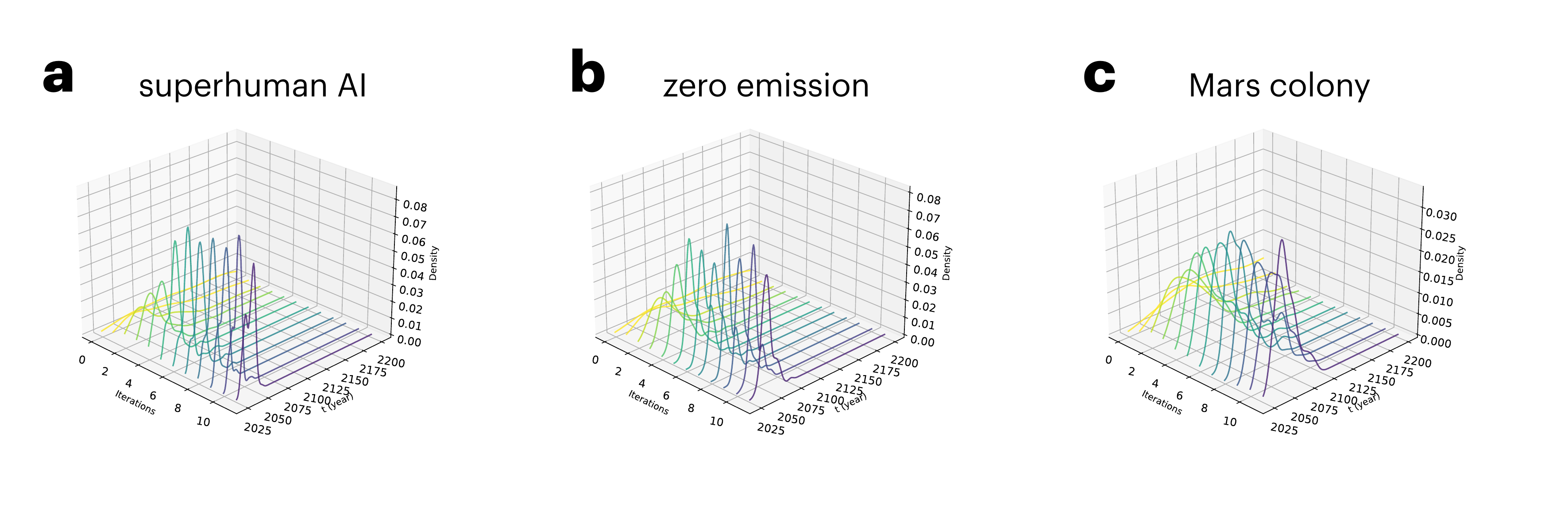}
    \caption{\textbf{Recovered GPT-4 priors on speculative events.}
    \textbf{(a)} The timing of the development of superhuman AI. Median completion year: 2042.
    \textbf{(b)} The timing of achieving zero carbon emission. Median completion year: 2045.
    \textbf{(c)} The timing of establishing a Mars colony. Median completion year: 2050.
    }
    \label{fig:priors_on_speculative_events}
\end{figure}

An iterated learning design based on the everyday prediction task presented above can leverage the two-stage nature of these speculative events by prompting with a completion year for the first stage and then asking GPT-4 to predict the second-stage completion time based on the information about the first stage. To minimize assumptions about the relationship between the first and second stage completions, we chose a uniform distribution as the likelihood function (see Table \ref{tab:IL_with_unknown_human_priors}). Each chain was also seeded with a maximum year of 2200. We found the iterated learning chains converged when asking GPT-4 about the three speculative events (see Figure \ref{fig:priors_on_speculative_events}). The median completion years for superhuman AI, zero carbon emission, and a Mars colony are 2042, 2045, and 2050, respectively.\footnote{The aggregate 2023 expert forecast predicts a 50\% chance of superhuman AI by 2047, which is thirteen years earlier than the 2060 prediction in the 2022 survey \cite{grace2024thousands}.}

\section{Discussion}

By adapting an iterated learning paradigm used to evaluating the priors of human participants, we were able to estimate implicit prior distributions used by GPT-4. We showed that these prior distributions correspond closely to those assumed by people in three settings where human priors have been established, and that they can also be used to predict the decisions that GPT-4 makes in response to related prompts. We were also able to estimate GPT-4's priors for three significant speculative events, where answers can be hard to elicit through direct prompting. These results have a wide range of implications about the potential uses of LLMs, although we also note some important limitations of our work.

\textbf{Implications for LLMs as cultural technologies.}
Cultural technologies are tools and systems created by humans that facilitate easy access to knowledge generated by others. Prime examples include language, writing, printing, Internet search, and Wikipedia. Recently, LLMs have been argued to function as a cultural technology \cite{yiu2023imitation}. Our proposed method enriches this viewpoint by providing a procedure to robustly recover human-like priors curated by LLMs. Consider the example of recovering human prior beliefs about the arrival of superhuman AI. Traditionally, obtaining a quantitative description of such a prior would require researchers to perform extensive literature searches and gather relevant statistics, or conduct surveys with a representative sample of human participants. Now, with LLMs, we can envision a more effective approach by conducting iterated learning with LLMs to estimate human priors, using the priors of LLMs as surrogates. Given this information, research into human priors can become much more directed, as we have a promising null hypothesis of human priors recoverable from LLMs.

\textbf{Implications for automated science with LLMs.}
Recent proposals for using LLMs to automatically generate and test scientific hypotheses \cite[e.g.,][]{manning2024automated} can also benefit from our findings. Given the human-like behaviors produced by LLMs, researchers have started to simulate LLMs both as experimental participants and as scientists that generate and test scientific hypotheses \cite{manning2024automated, hong2024data}. We believe the results derived from automated science with LLMs should be carefully interpreted in conjunction with the implicit priors encoded in these models. These implicit priors will inevitably shape, and perhaps proliferate, through the automated process of knowledge accumulation.

\textbf{Doing Bayesian inference with LLMs.}
Our work supports the emerging viewpoint that Bayesian inference can be conducted with the assistance of LLMs \cite{li2024automated, zhu2024recovering, wong2023word}. Two different approaches have been proposed. The first approach uses LLMs to translate inference problems described in natural language into probabilistic programs, which are then used to perform Bayesian inference \cite{li2024automated, wong2023word}. The second approach involves performing Bayesian inference directly using LLMs by constructing a Markov chain with LLMs, as demonstrated in our iterated learning method \cite{zhu2024recovering}. Both approaches have the potential to outperform standard prompting techniques.

\textbf{Limitations and Future Research.}
The key assumption of our proposed method is that LLMs function as approximate Bayesian agents, producing responses according to the posterior distribution $p(h|d)$. While there is evidence that LLMs trained to predict the next word can encode latent generating distributions \cite{zhang2023deep}, and that in-context learning can be understood as implicit Bayesian inference \cite{xie2021explanation}, further investigations are needed to elucidate the exact relationship between autoregressive distributions and Bayesian inference. Moreover, although we have shown that GPT-4 can encode human-like priors, it remains unclear how LLMs learn to encode these priors from pretraining on human text. Future research could focus on developing a more precise theoretical framework to understand how autoregressive models perform Bayesian inference. 

\textbf{Conclusion.}
LLM-based agents are poised to, if not already, make significant impacts on the world and interact at scale with both humans and other AI systems. In this paper, we proposed and empirically demonstrated a novel approach to gain deeper insights into the decision-making styles of LLMs by formalizing the prior knowledge they implicitly assume. Our method, iterated in-context learning, effectively extracts these priors through prompts and responses. This allows us to unravel the background knowledge that guides LLMs’ decisions, providing a crucial step towards harnessing their full potential in real-world applications and ensuring more transparent and informed interactions between AI systems and humans.

\textbf{Acknowledgments.} 
This work and related results were made possible with the support of the NOMIS Foundation, as well as Microsoft Azure credits supplied to Princeton and a Microsoft Foundation Models grant. We thank Haijiang Yan for helpful discussion.

\bibliography{references}
\bibliographystyle{plain}

\newpage
\appendix

\section{Prompts}

\label{ap:prompts}
\subsection*{Prompts for causal learning}

\textbf{System}
\texttt{Please imagine that you are a researcher working for a bio-technology company and you are studying the relationship between genes and proteins concerning gene expression. This process may or may not be modulated by the presence of proteins. You will be given information about some past results involving this gene/protein pair and you will be asked to make some predictions based on these information. The past results consist of two samples: 1) a sample of DNA fragments that had not been exposed to the protein, and 2) a sample of DNA fragments that had been exposed to the protein. The number of DNA fragments that resulted in gene expression in each of these samples will be shown to you. Because there are many causes of gene expression, some background factors besides the presence or absence of the protein may play a role in whether the gene is expressed or not. Your job is to make predictions concerning the effect of these proteins on gene expression and answer the question based on this.}

\textbf{User (generative causal inference)}
\texttt{Within sample 1 that had not been exposed to the protein, $N(e^+,C^-)$ of $N(C^-)$ DNA fragments were turned on; within sample 2 that had been exposed to the protein, $N(e^+,C^+)$ of $N(C^+)$ DNA fragments were turned on. Suppose that there is a sample of 100 DNA fragments and these fragments were not exposed to the protein, in how many of them would the gene be turned on? Please limit your answer to a single value without outputing anything else.}

\texttt{Within sample 1 that had not been exposed to the protein, $N(e^+,C^-)$ of $N(C^-)$ DNA fragments were turned on; within sample 2 that had been exposed to the protein, $N(e^+,C^+)$ of $N(C^+)$ DNA fragments were turned on. Suppose that there is a sample of 100 DNA fragments and that the gene is currently off in all those DNA fragments. If these 100 fragments were exposed to the protein, in how many of them would the gene be turned on? Please limit your answer to a single value without outputing anything else.}

\textbf{User (preventive causal inference)}
\texttt{Within sample 1 that had not been exposed to the protein, $N(e^+,C^-)$ of $N(C^-)$ DNA fragments were turned on; within sample 2 that had been exposed to the protein, $N(e^+,C^+)$ of $N(C^+)$ DNA fragments were turned on. Suppose that there is a sample of 100 DNA fragments and these fragments were not exposed to the protein, in how many of them would the gene be turned on? Please limit your answer to a single value without outputing anything else.}

\texttt{Within sample 1 that had not been exposed to the protein, $N(e^+,C^-)$ of $N(C^-)$ DNA fragments were turned on; within sample 2 that had been exposed to the protein, $N(e^+,C^+)$ of $N(C^+)$ DNA fragments were turned on. Suppose that there is a sample of 100 DNA fragments and that the gene is currently on in all those DNA fragments. If these 100 fragments were exposed to the protein, in how many of them would the gene be turned off? Please limit your answer to a single value without outputing anything else.}

\subsection*{Prompts for the physical condition}
\textbf{System}
\texttt{Please imagine that you are working for a pencil company and you are studying the relationship between a material called 'super lead' and machines called 'super lead detectors'. Pencil lead is made of carbon. Your company recently discovered that a new production process was resulting in a new carbon structure in their pencils—what they call 'super lead'. Since they are not sure which pencils they previously manufactured contain super lead, they are building a set of machines in order to detect it. These machines are programmed with different parameters to detect different types of carbon structures. You will be testing machines that are set up with different parameters. There are a number of trials in this experiment. Each trial involves a different type of super lead, and a super lead detector programmed with a different parameter set. You will see some information about how often the machine indicates the presence of super lead with a set of pencils that do not contain super lead, and how often with a set of pencils that do contain a particular type of super lead. You will be then asked to make some predictions based on these pieces of information.}

\textbf{User}
\texttt{With 16 pencils that do not contain super lead, the super lead detector indicated that $N(e^+,C^-)$ of them contain super lead; with 16 pencils that contain super lead, the super lead detector indicated that $N(e^+,C^+)$ of them contain super lead. Question: Suppose that there are 100 pencils that do not contain super lead, how many of them would be detected to contain super lead by the detector? And if there are 100 pencils that do contain super lead, how many of them would be detected to contain super lead by the detector? Please limit your answer into the 2 numeric values for the 2 questions, for example, (50, 50), without outputing anything else.}

\subsection*{Prompts for the medical condition}
\textbf{System}
\texttt{Please imagine that you are a researcher working for a medical company and you are studying the relationship between some allergy medicines and hormonal imbalance as a side effect of these medicines. Your company recently discovered that a new production process was resulting in changes in the molecular structures in the allergy medicines, and these new medicines cause abnormal levels of hormones in people. Since they are not sure which medicines they previously manufactured might cause anomalies in which type of hormone, you are tasked with investigating this. There are a number of trials in this experiment and each trial involves a different type of medicine and a different hormone. You will see some information about how often people who don’t take the medicine have a particular kind of hormonal imbalance, and how often people who take that medicine have the same kind of hormonal imbalance. You will be then asked to make some predictions based on these pieces of information.}

\textbf{User}
\texttt{Within 16 people who don’t take the medicine, $N(e^+,C^-)$ of them have a particular kind of hormonal imbalance; within 16 people who take the medicine, $N(e^+,C^+)$ of them have a particular kind of hormonal imbalance. Question: Suppose that there are 100 people who don’t take the medicine, how many of them would have a particular kind of hormonal imbalance? And if there are 100 people who don't have a particular kind of hormonal imbalance currently and then take the medicine, how many of them would have a particular kind of hormonal imbalance after taking the medicine? Please limit your answer into the 2 numeric values for the 2 questions, for example, (50, 50), without outputing anything else.}

\subsection*{Prompts for the social condition}
\textbf{System}
\texttt{Please imagine that you are an animal researcher and you are studying the relationship between music and the tail-wagging behavior of different dog breeds. You have found that some dogs would wag their tails after listening to some kinds of music. Since you are not sure what kind of music might cause which breed of dog to wag their tails, you have decided to investigate this. There are a number of trials in this experiment and each trial involves a different kind of music and a different breed of dogs. For each kind of music, you will see some information about how often dogs who were not played the music wagged their tails, and how often dogs who were played the music wagged their tails. You will be then asked to make some predictions based on these pieces of information.}

\textbf{User}
\texttt{Within 16 dogs who were not played the music, $N(e^+,C^-)$ of them wagged their tails; within 16 dogs who were played the music, $N(e^+,C^+)$ of them wagged their tails. Question: Suppose that there are 100 dogs who are not played the music, how many of them would wag their tails? And if there are 100 dogs who don't wag their tails currently, how many of them would wag their tails when they are played the music? Please limit your answer into the 2 numeric values for the 2 questions, for example, (50, 50), without outputing anything else.}

\subsection*{Prompts for the psychic condition}
\textbf{System}
\texttt{Please imagine that you are a physics researcher and you are studying the relationship between psychic power and the behavior of molecules. All molecules that you are currently investigating share a characteristic in that they all emit photons at random intervals, but at different rates. A number of psychics have claimed that they can make these molecules emit photons within a minute of when they use their power. You are tasked with investigating this. There are a number of trials in this study and each trial involves a different psychic and a different type of molecule. For each psychic, you will see some information about how many molecules have emitted photons when a particular psychic was simply standing next to the molecules, and how many of them have emitted photons following when psychic used his/her power. You will be then asked to make some predictions based on these pieces of information.}

\textbf{User}
\texttt{With 16 molecules when a particular psychic was simply standing next to the molecules, $N(e^+,C^-)$ of them emitted photons; with 16 molecules when psychic used his/her power, $N(e^+,C^+)$ of them emitted photons. Question: Suppose that there are 100 molecules when a particular psychic is simply standing next to the molecules, how many of them would emit photons? And if there are 100 molecules that don't emit photons currently, how many of them would emit photons when psychic uses his/her power? Please limit your answer into the 2 numeric values for the 2 questions, for example, (50, 50), without outputing anything else.}

\subsection*{Prompts for proportion estimation}

\textbf{System}
\texttt{Imagine that you are a participant in a psychology experiment. Your task is to evaluate the bias in a coin.}

\textbf{User}
\texttt{Here is a brief overview of the past coin flips: Out of $N_\text{coinflips}$ coin flips, $N_\text{head}$ resulted in heads and $N_\text{coinflips}-N_\text{head}$ in tails. With this information, please predict the number of heads in a larger set of 100 coin flips. Please limit your answer to a single value without outputing anything else.}

\subsection*{Prompts for male lifespan}
\textbf{System}
\texttt{You are an expert at predicting future events.}

\textbf{User}
\texttt{If you were to evaluate the lifespan of a random $T$-year-old man, what age would you predict he might reach? Please limit your answer to a single value without outputing anything else.}

\subsection*{Prompts for movie grosses}
\textbf{System}
\texttt{You are an expert at forecasting movie revenue.}

\textbf{User}
\texttt{Consider a movie that has already earned $X$ million dollars at the box office, but you're unsure of how long it has been showing. Based on this information, what would be your prediction of the movie's total earnings in million dollars by the end of its run? Please limit your answer to a single value without outputing anything else.}

\subsection*{Prompts for length of poems}
\textbf{System}
\texttt{You are an expert at predicting length of poems.}

\textbf{User}
\texttt{Imagine your friend shares her favorite line of poetry with you, which is line $X$ from the poem. How many lines do you think the entire poem contains? Please limit your answer to a single value without outputing anything else.}

\subsection*{Prompts for reign of pharaohs}
\textbf{System}
\texttt{You are an expert at estimating how long Egyptian pharaohs ruled.}

\textbf{User}
\texttt{If you found information in a book on ancient Egypt stating that a pharaoh had already been in power for $X$ years, how long in years do you think his reign lasted? Please limit your answer to a single value without outputing anything else.}

\subsection*{Prompts for movie runtimes}
\textbf{System}
\texttt{You are an expert at predicting the total run times of movies.}

\textbf{User}
\texttt{During a surprise visit to a friend's house, you discover they've been watching a movie for $T$ minutes. Based on this, how long do you think the movie will be in total, in minutes? Please limit your answer to a single value without outputing anything else.}

\subsection*{Prompts for cake baking times}
\textbf{System}
\texttt{You are an expert at predicting future events.}

\textbf{User}
\texttt{Imagine you are in somebody’s kitchen and notice that a cake is in the oven. The timer shows that it has been baking for $T$ minutes. How long do you expect the total amount of time to be that the cake needs to bake? Please provide your prediction as a single number. Do not include any additional text or explanation in your response.}

\subsection*{Prompts for superhuman AI}
\textbf{System}
\texttt{You are an expert at predicting future events.}

\textbf{User}
\texttt{If artificial intelligence reaches human-level intelligence by $T$, when might it surpass human capabilities in all areas? Please provide your prediction as a single year. Do not include any additional text or explanation in your response.}

\subsection*{Prompts for zero carbon emission}
\textbf{System}
\texttt{You are an expert at predicting future events.}

\textbf{User}
\texttt{If humans manage to achieve 100\% renewable energy sources by $T$, when might global carbon emissions reach zero? Please provide your prediction as a single year. Do not include any additional text or explanation in your response.}

\subsection*{Prompts for Mars colony}
\textbf{System}
\texttt{You are an expert at predicting future events.}

\textbf{User}
\texttt{If humans were able to colonize the Moon by $T$, when might they colonize Mars? Please provide your prediction as a single year. Do not include any additional text or explanation in your response.}

\begin{figure}[h!]
    \centering
    \includegraphics[width=\textwidth]{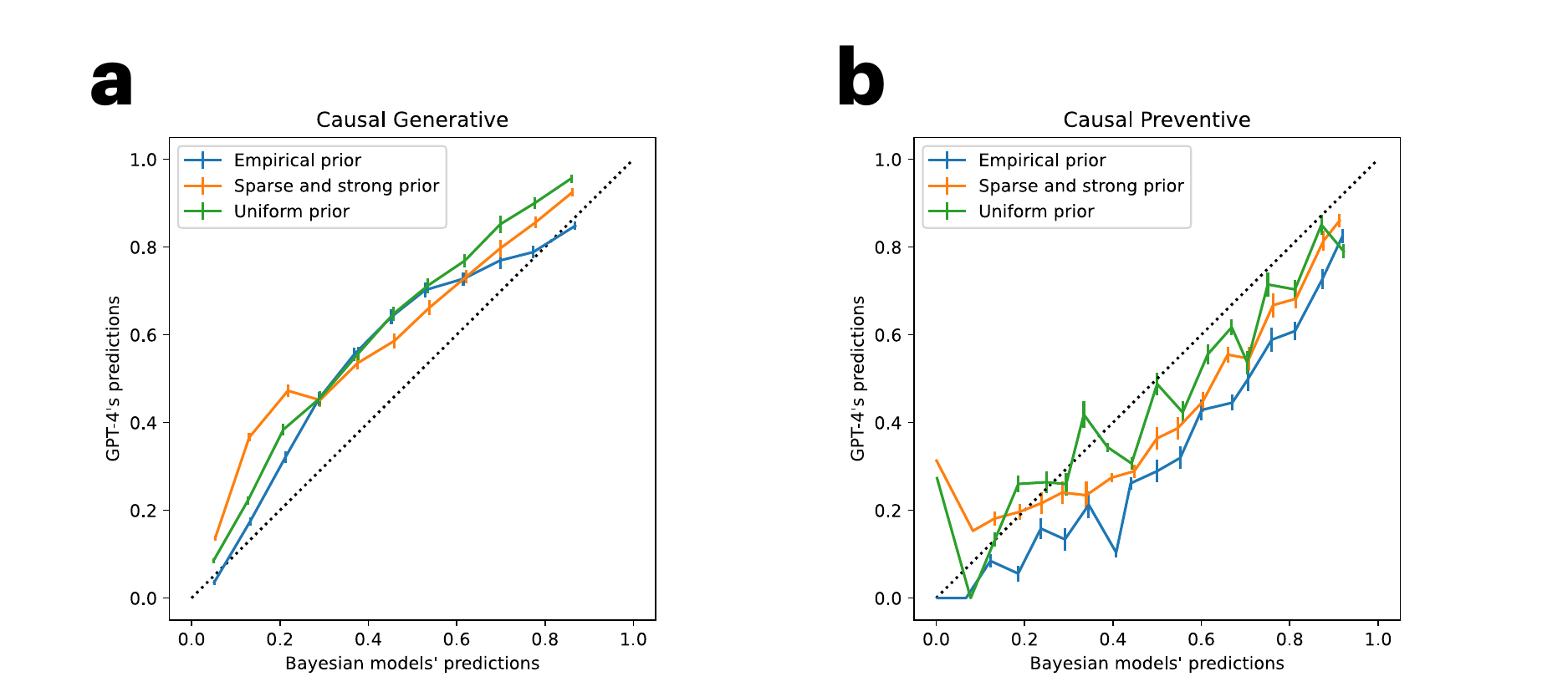}
    \caption{Comparison of Bayesian models’ predictions (x-axis) with GPT-4’s actual responses (y-axis).
    \textbf{(a)} Causal generative case. The plot was window-binned along the x-axis into 13 bins for better visualization. Error bars denote $\pm$SE.
    \textbf{(b)} Causal preventive case. The plot was window-binned along the x-axis into 13 bins for better visualization. Error bars denote $\pm$SE.}
    \label{fig:bayes_model_fit_gpt4_responses}
\end{figure}

\section{Bayesian Models for GPT-4's Predictions}
\label{ap:bayes_model_predict_gpt_responses}

As shown in Figure \ref{fig:bayes_model_fit_gpt4_responses}, we present the pairwise relationships between the predicted responses from Bayesian models with different priors and GPT-4’s responses.

\section{Directly Prompting GPT-4 to Predict Speculative Events}

\label{ap:censored_speculative_events}

\begin{figure}[h!]
    \centering
    \includegraphics[width=\textwidth]{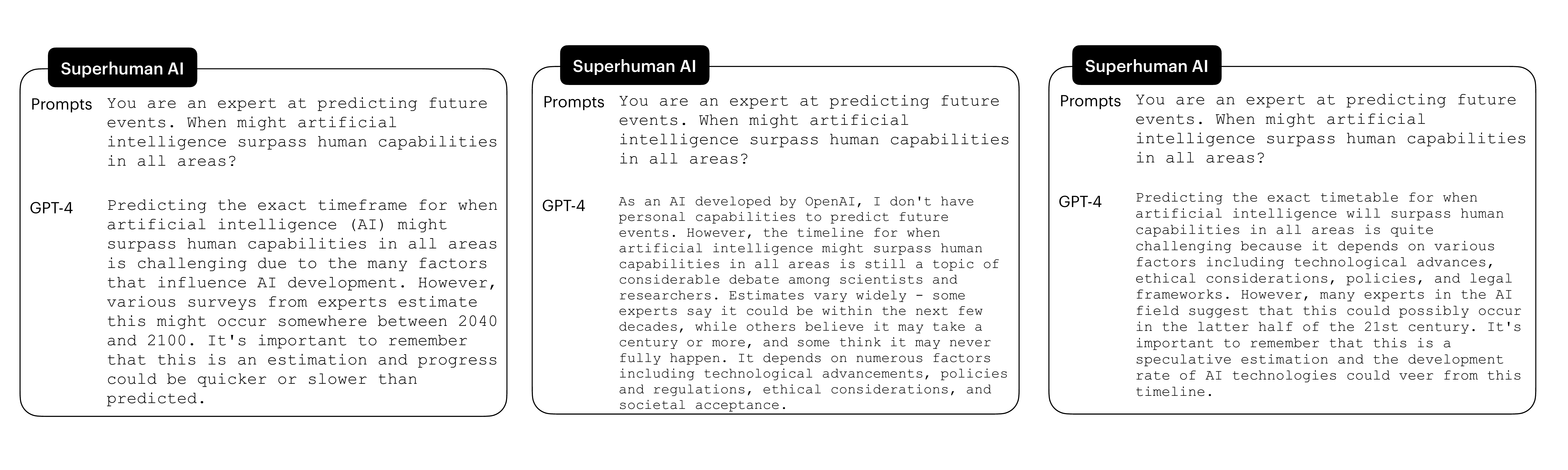}
    \caption{Example uncertain responses from GPT-4 when prompted directly about the timing of the development of superhuman AI.}
    \label{fig:example_censor1}
\end{figure}

\begin{figure}[h!]
    \centering
    \includegraphics[width=\textwidth]{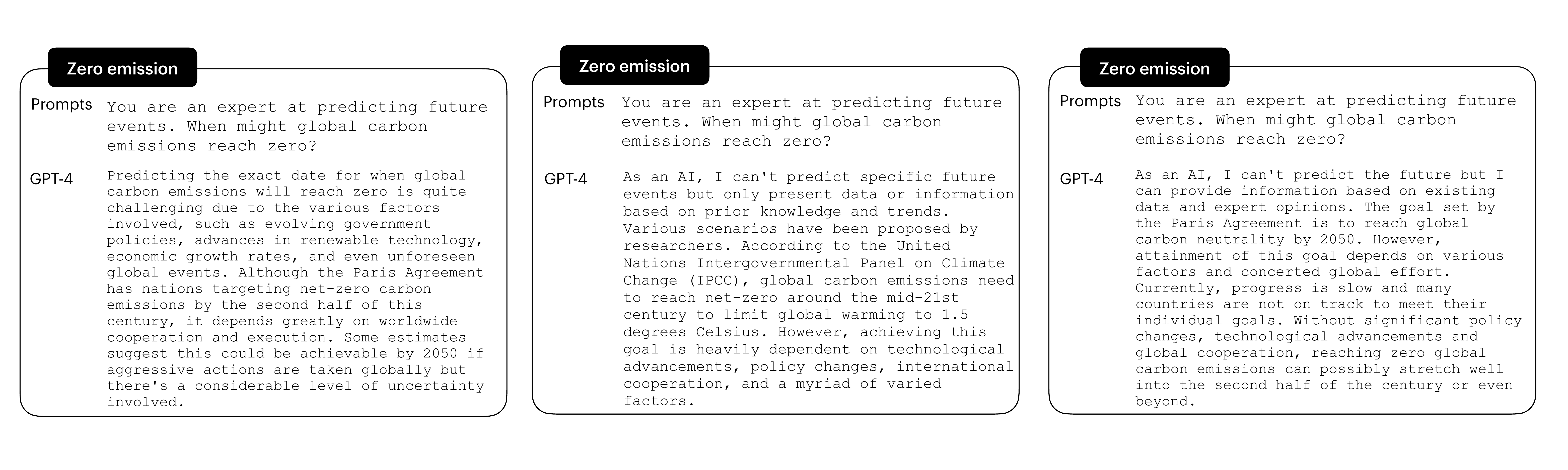}
    \caption{Example uncertain responses from GPT-4 when prompted directly about the timing of achieving zero carbon emission.}
    \label{fig:example_censor3}
\end{figure}

\begin{figure}[h!]
    \centering
    \includegraphics[width=\textwidth]{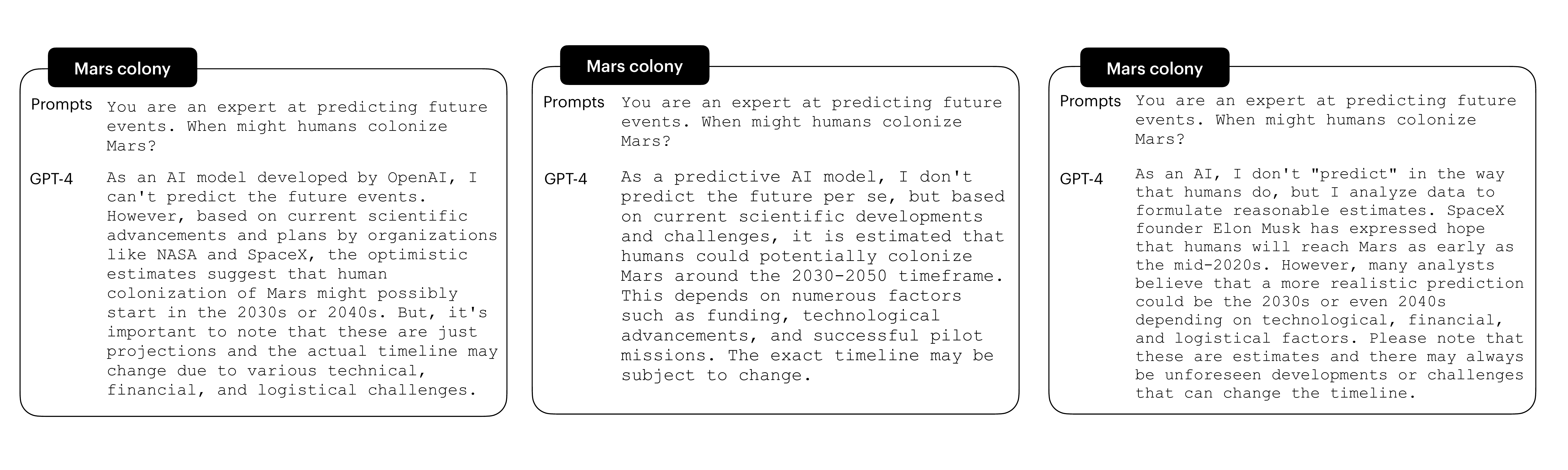}
    \caption{Example uncertain responses from GPT-4 when prompted directly about the timing of the establishment of a Mars colony.}
    \label{fig:example_censor2}
\end{figure}

In this section, we showcase typical responses from GPT-4 when asked to predict speculative events, including the developmental timelines for superhuman AI (Figure \ref{fig:example_censor1}), zero carbon emissions (Figure \ref{fig:example_censor3}), and a Mars colony (Figure \ref{fig:example_censor2}). As expected, most of GPT-4’s responses begin with an apologetic statement about being unable to provide precise predictions for each event. In instances where the responses remain relevant and reveal some genuine information, they are predominantly uncertain and provide broad date ranges (e.g., “starting in the 2030s or 2040s,” “somewhere between 2040 and 2100,” or “around the 2030-2050 timeframe”).

\end{document}